\documentclass[10pt,twocolumn,letterpaper]{article}

\usepackage[pagenumbers]{cvpr} 

\usepackage[utf8]{inputenc} 
\usepackage[T1]{fontenc}    
\usepackage{latexsym}
\usepackage{microtype}
\usepackage{booktabs}
\usepackage{graphicx}
\graphicspath{{figures/}}
\usepackage{enumitem}
\usepackage{mathabx,amsbsy,mathtools,etoolbox,mathbbol}
\usepackage{algorithmic}
\usepackage[linesnumbered,ruled,vlined]{algorithm2e}
\usepackage{acronym}
\usepackage{balance}
\usepackage{xspace}
\usepackage{setspace}
\usepackage[skip=3pt,font=small]{subcaption}
\usepackage[skip=3pt,font=small]{caption}
\usepackage{tabularx,colortbl,multirow,array,makecell}
\usepackage{overpic,wrapfig}
\usepackage{nicefrac}
\usepackage[misc]{ifsym}
\usepackage{siunitx} 
\usepackage{soul} 
\usepackage{pifont}
\usepackage{relsize}
\usepackage{fontawesome5}
\usepackage[export]{adjustbox} 

\makeatletter
\DeclareRobustCommand\onedot{\futurelet\@let@token\@onedot}
\def\@onedot{\ifx\@let@token.\else.\null\fi\xspace}
\def\eg{\emph{e.g}\onedot} 

\def\ie{\emph{i.e}\onedot}

\def\etc{\emph{etc}\onedot} 
\def\vvss{\emph{vs}\onedot}

\makeatother
\newcommand{\sota}{\text{state-of-the-art}\xspace}
\newcommand{\supp}{\textit{supplementary}\xspace}

\newcommand{\cmark}{\ding{51}}
\newcommand{\xmark}{\ding{55}}

\newcommand{\model}{\textsc{MPEC}\xspace}

\acrodef{3dvl}[3D-VL]{3D vision-language}
\acrodef{2dvl}[2D-VL]{2D vision-language}
\acrodef{eai}[EAI]{embodied AI}
\acrodef{sota}[SOTA]{state-of-the-art}
\newcommand{\sotas}{\text{state-of-the-arts}\xspace}
\acrodef{ovsem}[OV-SemSeg]{open-vocabulary semantic segmentation}

\acrodef{pku}[PKU]{Peking University}
\acrodef{thu}[THU]{Tsinghua University}
\acrodef{bigai}[BIGAI]{Beijing Institute of General Artificial Intelligence}

\def\pcd{{\bm{\mathcal{P}}}}
\def\cossim{d_{\text{cos}}}

\definecolor{phcolor}{RGB}{188, 188, 188}
\usepackage{lipsum}

\newcommand\blfootnote[1]{%
  \begingroup
  \renewcommand\thefootnote{}\footnote{#1}%
  \addtocounter{footnote}{-1}%
  \endgroup
}

\usepackage{amsmath,amsfonts,bm}

\def\eqref#1{equation~\ref{#1}}

\def\1{\bm{1}}

\def\vf{{\bm{f}}}

\def\vs{{\bm{s}}}
\def\vt{{\bm{t}}}

\def\mC{{\bm{C}}}

\def\mE{{\bm{E}}}
\def\mF{{\bm{F}}}
\def\mG{{\bm{G}}}

\def\mM{{\bm{M}}}

\def\mP{{\bm{P}}}

\def\mT{{\bm{T}}}

\DeclareMathAlphabet{\mathsfit}{\encodingdefault}{\sfdefault}{m}{sl}
\SetMathAlphabet{\mathsfit}{bold}{\encodingdefault}{\sfdefault}{bx}{n}

\def\gP{{\mathcal{P}}}

\def\sR{{\mathbb{R}}}

\definecolor{cvprblue}{rgb}{0.21,0.49,0.74}
\usepackage[pagebackref,breaklinks,colorlinks,allcolors=cvprblue]{hyperref}

\def\paperID{5504} 
\def\confName{CVPR}
\def\confYear{2025}

\title{Masked Point-Entity Contrast for Open-Vocabulary 3D Scene Understanding}

\author{
    Yan Wang$^{1*}$ \qquad Baoxiong Jia$^{1*}$ \qquad Ziyu Zhu$^{1,2}$ \qquad Siyuan Huang$^{1}$ \\
    $^1$State Key Laboratory of General Artificial Intelligence, BIGAI \\
    $^2$Tsinghua University \\
    \url{https://mpec-3d.github.io}
}

\begin{document}
\twocolumn[{
\renewcommand\twocolumn[1][]{#1}%
\maketitle
\begin{center}
    \centering
    \vspace{-15pt}
    \captionsetup{type=figure}
    \includegraphics[width=\linewidth]{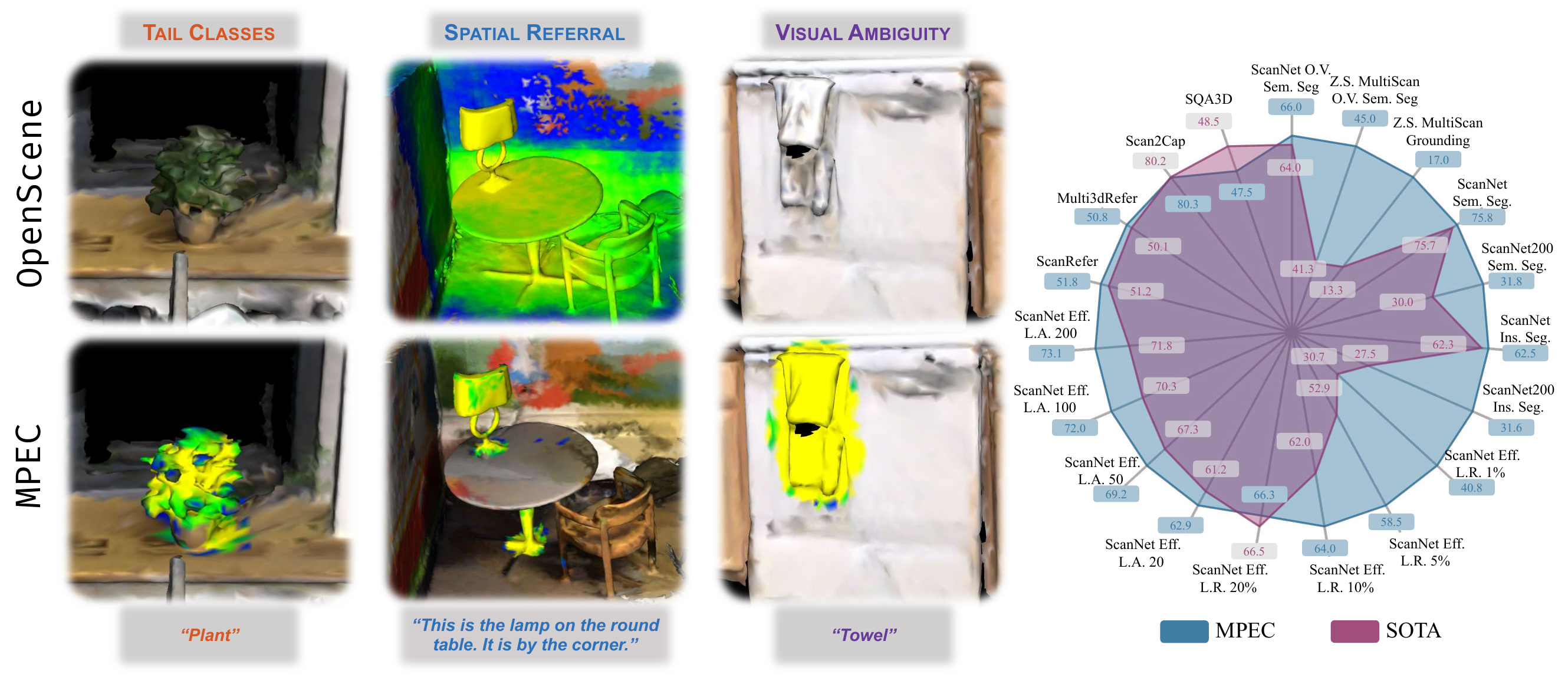}
    \caption{\textbf{Qualitative and Quantitative Analysis of \model Results.} \model achieves \sota on open-vocabulary 3D semantic segmentation. Compared with OpenScene~\cite{peng2023openscene}, \model is more robust to tail classes, visual ambiguity, and detailed descriptions, \eg, spatial referral. The radar chart highlights the performance advantages of \model across various 3D scene understanding tasks.}
    \label{fig:teaser}
\end{center}%
}]

\blfootnote{$^*$ indicates equal contribution.}
  
\begin{abstract}
Open-vocabulary 3D scene understanding is pivotal for enhancing physical intelligence, as it enables embodied agents to interpret and interact dynamically within real-world environments. This paper introduces \textbf{\textbf{\model}}, a novel \textbf{M}asked \textbf{P}oint-\textbf{E}ntity \textbf{C}ontrastive learning method for open-vocabulary 3D semantic segmentation that leverages both 3D entity-language alignment and point-entity consistency across different point cloud views to foster entity-specific feature representations. \model improves semantic discrimination and enhances the differentiation of unique instances, achieving \sota results on ScanNet for open-vocabulary 3D semantic segmentation and demonstrating superior zero-shot scene understanding capabilities. Extensive fine-tuning experiments on 8 datasets, spanning from low-level perception to high-level reasoning tasks, showcase the potential of learned 3D features, driving consistent performance gains across varied 3D scene understanding tasks.
\end{abstract}
\section{Introduction}
\label{sec:intro}

Open-vocabulary 3D scene understanding~\cite{peng2023openscene} is an emerging research topic with significant implications for embodied AI~\cite{ahn2022can, huang2023embodied, hong20233d, zhi2024closed, firoozi2023foundation, o2023open, yu2025metascenes}. This task requires the model to act beyond recognizing only predefined 3D object categories and concepts seen during training, enabling it to identify flexible concepts based on instructions while preserving sufficient spatial or geometry information to support tasks like reconstruction~\cite{ni2025dprecon, lu2025movis, liu2025building, ni2024phyrecon} and manipulation~\cite{gu2024conceptgraphs,zhi2024closed, li2025maniptrans}.

However, developing scene representations that possess both flexible semantic understanding capability and accurate spatial and geometric information presents non-trivial challenges. Due to the lack of available paired \ac{3dvl} data, recent methods leverage 2D foundation models~\cite{kirillov2023segment, ghiasi2022scaling, li2022language, radford2021learning} either to generate data~\cite{ding2023pla,yang2024regionplc,jiang2024open} for 3D-language alignment training or to provide semantically aligned 2D representations from individual images which are then merged into global 3D representations~\cite{xu2023sampro3d, yang2023sam3d} or distilled into trainable 3D encoders~\cite{peng2023openscene}. However, such pipelines can struggle to preserve 3D spatial information, primarily due to the limited field of view in individual images, which hinders the capture of global spatial relationships between objects, and the challenges in maintaining multi-view consistency, especially by semantic representations rather than pixel-level information. Considering the intrinsic uniqueness of individual entities in 3D scenes, particularly for tasks requiring spatial reasoning in \ac{3dvl} and embodied AI, we emphasize the need for representations that not only align with language but also distinguish between instances within the scene.

Previous methods for incorporating spatial and geometric information without supervision have largely focused on investigating the inherent correlations in point clouds, with much of the work centered on unsupervised and self-supervised representation learning~\cite{yu2022point}, particularly via contrastive learning approaches~\cite{xie2020pointcontrast, hou2021exploring, wu2023masked, wang2024groupcontrast}. These methods typically create positive and negative point cloud pairs through various strategies to form the concept of neighbors~\cite{wu2023masked}, regions~\cite{hou2021exploring}, and semantic prototypes~\cite{wang2024groupcontrast}. However, none of these approaches are effectively suited for aligning with language in complex 3D scenes, as they tend to be either overly fragmented or coarse-grained and fail to capture the concept of objects within the scene.

In this paper, we address the aforementioned challenges \textit{by learning 3D representations that preserve both flexible semantic understanding and entity-specific geometric and spatial information.} To achieve this, we propose \model, \textit{a simple yet effective entity-level contrastive framework for training open-vocabulary 3D scene understanding models.} Our approach bridges efforts from open-vocabulary scene understanding and unsupervised scene representation learning~\cite{wu2023masked, wang2024groupcontrast}, while also leveraging recent advancements in \ac{3dvl} scaling~\cite{wang2024embodiedscan,jia2024sceneverse,yang20243d,yu2025metascenes}. Methodologically, we apply augmentations~\cite{wu2023masked} to generate multiple views of a scene. We generate 3D object entity masks with off-the-shelf object models~\cite{takmaz2023openmask3d,schult2023mask3d}, and form contrastive pairs between different views, enforcing feature similarity within the same entities across views and dissimilarity across different ones. Finally, we align 3D point features merged from different views with language generated by foundation models~\cite{yang2024regionplc,jiang2024open,jia2024sceneverse} on diverse scenes through contrastive alignment. 

\model achieves \sota open-vocabulary 3D semantic segmentation results with 66.0\% foreground mIoU and 81.3\% foreground mAcc on ScanNet~\cite{dai2017scannet}. The trained 3D encoder also demonstrates strong generalization across various downstream tasks after fine-tuning, from low-level perception~\cite{graham20183d, jiang2020pointgroup} to high-level reasoning tasks~\cite{chen2020scanrefer, achlioptas2020referit3d, ma2022sqa3d, chen2021scan2cap, zhang2023multi3drefer}. 
Notably, \model especially boosts the performance of ScanNet Data Efficient Benchmark~\cite{hou2021exploring}, \textit{improving mIoU from 30.7\%~\cite{wang2024groupcontrast} to 40.8\% with just 1\% of the scenes trained}. We also provide an empirical analysis of the framework, breaking down each module's contribution to the overall performance. In addition, we conduct extensive experiments to assess how different scene sources and language types affect performance. In summary, our contributions are:

\begin{itemize}[leftmargin=*,noitemsep,nolistsep]
    \item We present \model, a novel contrastive framework with both point-to-entity and entity-to-language alignments, achieving new \sotas in open-vocabulary 3D scene understanding tasks.
    \item Through extensive experiments on 7 tasks over 8 datasets, we demonstrate that \model achieves \sota results in both zero-shot and fine-tuned settings, particularly excelling in data-scarce scenarios.
    \item We analyze how model architectures and data sources impact model performance, highlighting cross-view contrast and data domain selection as key factors for contrastive learning, offering new insights for future research.
\end{itemize}
\section{Related Work}
\label{sec:related_work}

\paragraph{Open-Vocabulary 3D Scene Understanding}
Open-vocabulary 3D scene understanding typically involves two signature tasks: open-vocabulary 3D semantic segmentation~\cite{ding2023pla, peng2023openscene, yang2024regionplc, jiang2024open, chen2023clip2scene, wang2024open, li2024dense} and open-vocabulary 3D instance segmentation~\cite{takmaz2023openmask3d, nguyen2024open3dis, huang2025openins3d}. Both tasks aim to predict 3D segmentation masks for class labels or object instances using natural language descriptions, without requiring prior model training on specific object categories. Despite significant advancements in open-vocabulary understanding within the 2D domain~\cite{jia2021scaling, radford2021learning, zhou2022detecting, cho2024cat, wu2023clipself, xie2024sed, ghiasi2022scaling, li2022language}, achieving open-vocabulary 3D scene understanding from point clouds~\cite{takmaz2023openmask3d, yan2024maskclustering, peng2023openscene, ding2023pla, yang2024regionplc, jiang2024open, jia2024sceneverse, huang2025unveiling} remains challenging due to the scarcity of large-scale training data.

Existing approaches leverage 2D foundation models~\cite{kirillov2023segment, radford2021learning, ghiasi2022scaling, li2022language} to bridge the gap between language and 3D point cloud data. With off-the-shelf 3D segmentation models~\cite{schult2023mask3d,lu2023ovir} or back-projection from 2D foundation models~\cite{kirillov2023segment,zhou2022detecting}, existing methods learn point-language alignment for open-vocabulary semantic~\cite{peng2023openscene, ding2023pla,yang2024regionplc,jiang2024open} and instance segmentation~\cite{takmaz2023openmask3d,huang2025openins3d}. These methods align 3D point clouds and language either by directly distilling language and image features from \ac{2dvl} models into 3D learnable backbones or by contrastively aligning point and language features, similar to approaches in 2D~\cite{radford2021learning}. However, a key limitation of these models is the insufficient incorporation of 3D information during model learning, which restricts their performance on a wide range of \ac{3dvl} tasks. For instance, in OpenScene~\cite{peng2023openscene}, as illustrated in~\cref{fig:teaser}, the model struggles with 2D visual ambiguities (\eg, objects with similar colors) and faces challenges in accurately identifying specific object instances. This issue arises primarily because the \ac{2dvl} model used to train the 3D scene encoder was based on category-level text descriptions, which lack spatial and 3D geometric context, thereby limiting the model's understanding of the 3D structure of objects. Therefore, we emphasize the crucial role of 3D information from the scene representation itself in open-vocabulary scene understanding tasks.

\vspace{-3mm}
\paragraph{Pre-training for 3D Point Clouds}
Motivated by the successful application of the ``pre-train then fine-tune'' paradigm in 2D vision research, there has been growing interest in 3D point cloud representation learning~\cite{xie2020pointcontrast, huang2021spatio, hou2021exploring, wu2023masked, wang2024groupcontrast, yu2022point, zhang2021self, min2023occupancy, tian2023geomae, yang2023gd, wei2023t}, demonstrating positive impacts on downstream tasks. These approaches can be categorized into two paradigms: mask modeling and contrastive learning. Mask modeling methods~\cite{yu2022point, zhang2021self, min2023occupancy, tian2023geomae, yang2023gd, wei2023t} focus on reconstructing the coordinates or features of masked points. In contrast, contrastive learning methods~\cite{xie2020pointcontrast, huang2021spatio, hou2021exploring, wang2024groupcontrast} create positive and negative pairs within 3D point clouds for discriminative learning. To enhance the learning of contrastive-based methods, MSC~\cite{wu2023masked} incorporates masked point modeling as additional reconstructive supervision. GroupContrast~\cite{wang2024groupcontrast} builds on this by introducing semantic prototypes for discovering local neighborhoods. However, these methods still lack a concept of objects, making them inapplicable for aligning with language.

Meanwhile, recent research has shown a growing interest in vision-language pre-training for 3D point cloud understanding. Methods like 3D-VisTA~\cite{zhu20233d}, 3DVLP~\cite{3D-VLP}, and SynVL3D~\cite{Syn3DVL} employ unsupervised pre-training strategies such as masked modeling and contrastive learning~\cite{jia2024sceneverse}, while other approaches utilize multi-task training for vision-language alignment~\cite{zhu2024unifying, chen2023unit3d, cai20223djcg}. With the emergence of 3D Large Language Model (3D-LLM)~\cite{3d-llm,3d-llm-survey, robin3d, Multiply, 3d-vla}, recent efforts like LEO~\cite{huang2023embodied}, MiniGPT-3D~\cite{mini-GPT3D} and LLaVA-3D~\cite{llava-3d} have focused on aligning object-level or voxel-level features with LLMs through explicit alignment training stage~\cite{liu2023llava, 3d-llm}. These studies demonstrate that improved vision-language alignment leads to enhanced performance on downstream tasks~\cite{scene-llm, qi2025shapellm, chat-scene, pointbind, huang2023embodied, huang2025unveiling}. However, a critical gap remains in the lack of a robust, open-vocabulary 3D encoder capable of directly processing scene-level point clouds ~\cite{zhu2024unifying, huang2023embodied,llava-3d, huang2025unveiling}. Our work addresses this need by introducing a powerful 3D encoder, which has the potential to further advance the capabilities of 3D-LLM.
\section{Method}
\label{sec:method}

\begin{figure*}[htbp]
    \centering
    \includegraphics[width=\linewidth]{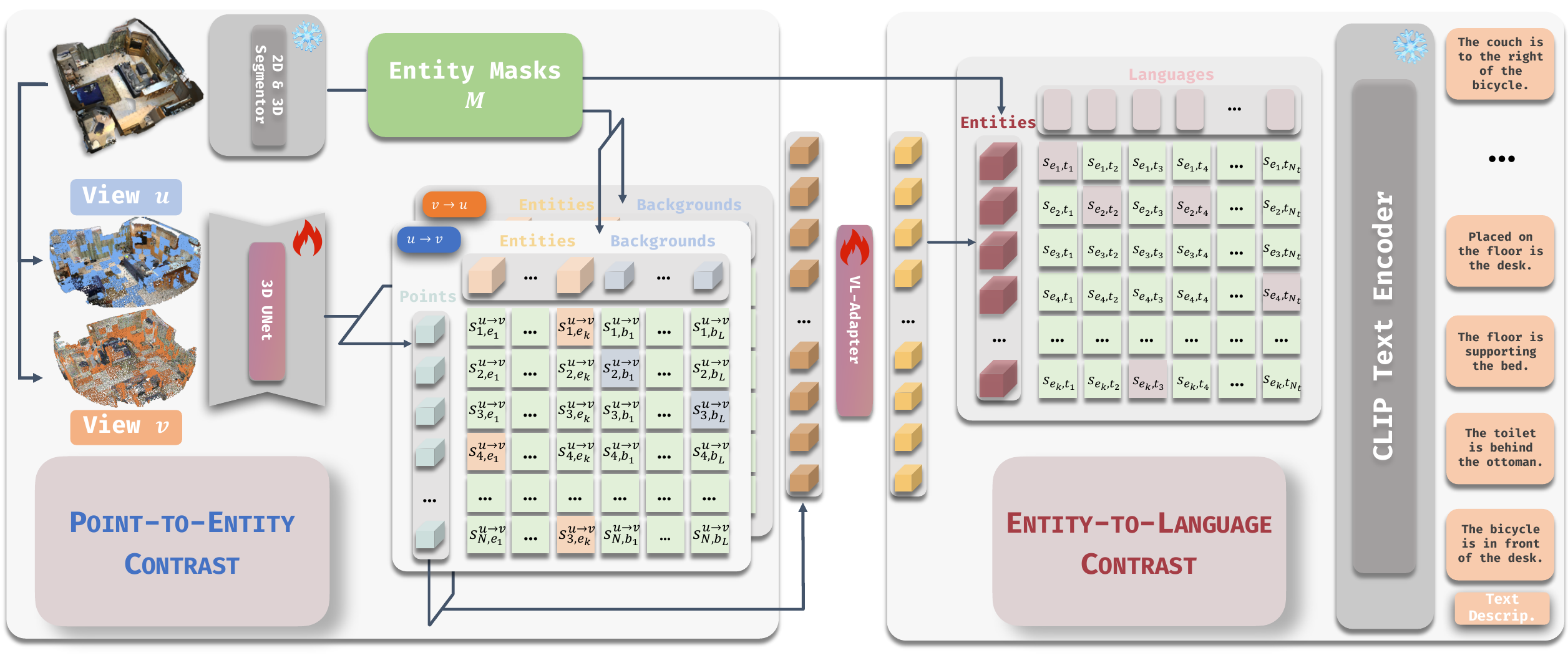}
    \caption{\textbf{The Overall Pipeline of \model.} Given a 3D point cloud as input, we predict entity mask proposals and generate text descriptions for each entity. Then different views of the scene are randomly masked and we replace the masked point features with learnable embeddings. A 3D UNet then extracts per-point features. Guided by the entity masks, cross-view point-to-entity contrastive learning is conducted to enforce cross-view consistency and distinguishment across different entities. Then we merge the point features from two views and align the feature dimension of points with text features extracted by CLIP with a VL-Adapter. Finally, entity-to-language contrastive learning is performed for open-vocabulary 3D scene understanding.}
    \vspace{-3mm}
    \label{fig:model}
\end{figure*}

In this section, we elaborate on the proposed masked point-entity contrastive learning framework for open-vocabulary 3D scene understanding. The overall architecture is presented in~\cref{fig:model}. In essence, our model aims to learn 3D point features $\mF_P\in\sR^{N\times D}$ that maintain 3D information of 3D scenes, a vision-language (VL-) adapter $\mathrm{Proj}_{\text{VL}}(\cdot)$ that maps these 3D features into the semantic space which is aligned with language features $\mF_{T}$ extracted from a pre-trained text encoder. The input of our framework mainly consists of three components: \begin{enumerate*}[label=\arabic*)]
    \item 3D scene point cloud $\pcd = \left( \mP, \mC \right)$, where $\mP \in \sR^{N \times 3}$ and $\mC \in \sR^{N \times 3}$ represents spatial coordinates and colors of the points.
    \item $K$ 3D entity mask proposals obtained with off-the-shelf methods which are merged into $\mM \in \{ \texttt{BG}, 1, \cdots, K\}^{N}$ where \texttt{BG} denotes the background.
    \item Text descriptions $\mT = \{t_i\}_{i=1}^{N_T}$ including captions and referrals, which could be generated for each entity proposal by template-based methods or foundation models.
\end{enumerate*} %

\subsection{Point-to-Entity Alignment}
\label{method:e2e} %
\paragraph{Cross-view Feature Extraction} We start by extracting separate point features from different augmented views of the scene for contrastive representation learning. Specifically, given a 3D point cloud $\pcd$, two distinct augmented views $\pcd_{u}$ and $\pcd_{v}$ are generated, along with their transformed entity masks $\mM_{u}$ and $\mM_{v}$. Following the practice of previous works~\cite{xie2022simmim, he2022masked, wu2023masked}, we divide the unioned set of point clouds into non-overlapping grids based on the original spatial locations. Then two sets of non-overlapping grid masks $\mG_{u}$ and $\mG_{v}$ are randomly generated and applied to the corresponding
views. We use the same mask token embedding $\vt\in\sR^{3}$ for the color of all masked points. We then feed the masked point clouds of separate views into the 3D U-Net encoder $\bm{\mathcal{E}}_{\text{pcd}}$ to extract per-point embeddings:
\begin{equation} \label{eq:extract_point_feats}
    \begin{aligned}
        \mF_{l} = \bm{\mathcal{E}}_{\text{pcd}}((1 - \mG_{l})\pcd_{l} + \mG_{l}\mT_{l}), && l\in\{u, v\},
    \end{aligned}
\end{equation}
where $\mT_{u}$ and $\mT_{v}$ are the expansions of the learnable embedding vector to fit the dimension with point cloud across different views, respectively.

\vspace{-3mm}
\paragraph{Point-to-Entity Contrastive Learning} After obtaining the point features from two views, we perform point-to-entity contrastive learning to incorporate entity information into our learned scene representations. Specifically, we calculate the cosine similarity matrix $\vs^{u \to v}$ between all pairs of matched points by computing the similarities of point $i$ from view $u$ to points in view $v$ belonging to entities and background separately. Let $\cossim(\vf_1,\vf_2) = \frac{\vf_1^T\cdot \vf_2}{\|\vf_1\|\cdot\|\vf_2\|}$ to denote the cosine similarity function, for points $\mE_k^v = \{j \mid \mM_v[j] = e_k\}$ that belong to entity $e_k$ in view $v$, we have:
\begin{equation}
    \vs^{u \to v}_{i,e_k} = \frac{1}{|\mE_k|} \sum_{j\in \mE_k} s^{u \to v}_{i,j} = \frac{1}{|\mE_k|} \sum_{j\in\mE_k} \cossim(\mF_u^i, \mF_v^j).
\end{equation}
Meanwhile, for the remaining $L$ background points $\{b_1, b_2, \cdots, b_L\}$ in view $v$, we compute the similarity between point $i$ to the each $b_j$ in view $v$ with cosine similarity $\vs_{i,b_j}^{u\to v} = \cossim(\mF_u^i, \mF_v^{b_j})$. After these calculations, we obtain the entity-aware point similarities $\vs_i^{u\to v}$ for point $i$ by
\begin{equation}
    \scalebox{0.95}{$\vs_i^{u\to v} = \mathrm{Concat}\left(\vs_{i,e_1}^{u\to v}, \cdots, \vs_{i, e_K}^{u\to v}, \vs_{i,b_1}^{u\to v}, \cdots, \vs_{i,b_L}^{u\to v}\right)$},
\end{equation}
and add point-entity contrastive supervision depending on whether point $i$ belongs to an entity or background in view $u$. If point $i$ belongs to an entity, then we have
\begin{equation}
    \ell_{i}^{u \to v} = -\log \frac{\exp(\vs^{u \to v}_{i,\bar{e}} / \tau)}{\sum_r \exp(\vs^{u \to v}_{i,r} / \tau)}, 
\end{equation} where $\bar{e}$ is the entity in view $v$ that point $i$ belongs to. If point $i$ belongs to the background, then we use
\begin{equation}
    \ell_{i}^{u \to v} = -\log \frac{\exp(\vs^{u \to v}_{i,\bar{b}} / \tau)}{\sum_r \exp(\vs^{u \to v}_{i,r} / \tau)}, 
\end{equation}
to align point $i$ with its corresponding background point $\bar{b}$ in view $v$ with a temperature factor $\tau$. We can similarly compute the similarity from view $v$ to $u$ and obtain the final point-entity contrastive loss:
\begin{equation} \label{loss:e2e}
\begin{aligned}
    \mathcal{L}_\text{p2e} = \frac{1}{2}\left(\frac{1}{N}\sum_{i}\ell_i^{u\to v} + \frac{1}{N}\sum_{j}\ell_j^{v\to u}\right).
\end{aligned}
\end{equation}

\subsection{Entity-to-Language Alignment}
\label{method:e2l}
To achieve open-vocabulary 3D scene understanding, we align the 3D point features with language features. First, we merge the 3D point features from both views to generate the final point features $\mF_\gP$. Next, we feed the text descriptions of 3D entities to a frozen text encoder $\bm{\mathcal{E}}_{\text{txt}}$ to extract language features $\mF_{T}$. Then we perform entity-to-language contrastive learning to align the 3D point features with their corresponding language features.

\vspace{-3mm}
\paragraph{Point Features Merging} In~\cref{method:e2e}, we extract point features $\mF_{u}$ and $\mF_{v}$ for different augmented views of the input point cloud. To fully utilize the advantages of both views' features and further align with language features, we merge $\mF_{u}$ and $\mF_{v}$ to obtain point features $\mF_\gP$ by taking the mean feature of matched points in the two views. We then feed the merged features to the VL-adapter $\mathrm{Proj}_{\text{VL}}(\cdot)$ to align the dimension of point and language features:
\begin{equation}
\begin{aligned}
    \mF_\gP = \mathrm{Mean}(\mF_u, \mF_v), && \mF_{\text{VL}} = \mathrm{Proj}_{\text{VL}}(\mF_\gP).
\end{aligned}
\end{equation}

\vspace{-3mm}
\paragraph{\textbf{Language Feature Extraction}}
Given various text descriptions $\mT$, \eg, object captions $\mT_\text{cap}$ and referrals $\mT_\text{ref}$ of different 3D entities in the scene, we random sample $N_\text{T}$ descriptions and extract their embeddings using the pre-trained CLIP text encoder $\bm{\mathcal{E}}_{\text{txt}}$ of CLIP~\cite{radford2021learning}:
\begin{equation}
\mF_{T} = \bm{\mathcal{E}}_{\text{txt}}(\mT) \in \sR^{N_\text{T}\times D}.
\end{equation}

\vspace{-3mm}
\paragraph{Entity-to-Language Contrastive Learning}
After obtaining the adapted 3D point features $\mF_{\text{VL}}$ and entity text description features $\mF_{T}$, we conduct entity-to-language contrastive learning to align their representations. 
Analogous to~\cref{method:e2e}, we employ a point-discriminative loss to enforce consistency among matched point-language pairs, while contrasting unmatched pairs.
Specifically, given the language features $\mF_{T}$ and adapted vision features $\mF_{\text{VL}}$, we first compute the cosine similarity matrix $\vs^{t\to e}$. For language description $i$, we compute its similarity to all points in an entity $e_k$ by
\begin{equation}
\begin{aligned}
    \vs^{t \to e}_{i,e_k}= \frac{1}{|\mE_k|} \sum_{j\in \mE_k}\cossim(\mF_T^i, \mF_{\text{VL}}^j), 
\end{aligned}
\end{equation}
where $\mE_k = \{j | \mM[j] = e_k\}$ denotes the set of points that belongs to entity $e_k$. We concatenate text-to-entity similarities $\vs_i^{t\to e}$ between text $i$ and all entity features derived from $\mF_{\text{VL}}$ by
\begin{equation}
\begin{aligned}
    \vs_i^{t \to e} = \mathrm{Concat}(\vs^{t \to p}_{i,e_1}, \vs^{t \to p}_{i,e_2}, \cdots, \vs^{t \to e}_{i,e_K}). 
\end{aligned}
\end{equation}

Let $\bar{e}$ denote the entity that aligns with text description $i$, we compute the text-to-entity contrastive loss with temperature factor $\tau$ as:
\begin{equation}
\begin{aligned}
    \ell^{t \to e}_i = -\log \frac{\exp(\vs^{t \to e}_{i,\bar{e}} / \tau)}{\sum_k \exp(\vs^{t \to e}_{i, k} / \tau)}.
\end{aligned}
\end{equation} 

Similarly, we can compute the entity-to-text similarity $\vs^{e \to t}$. However, in contrast to the direct symmetry between point-to-entity and entity-to-point alignment in~\cref{method:e2e}, the entity-to-text loss $\ell^{e\to t}$ needs adjustments as different descriptions can refer to the same object entity in the scene and thus disable the application of the cross-entropy loss. Therefore, we replace the loss with a binary cross-entropy loss when applying the entity-to-language supervision:

\begin{equation}
\scalebox{0.9}{%
$
\begin{aligned}
\ell^{e \to t}_{k} = - \frac{1}{N_{T}}\sum_{t\in\mT} & \left[ \mathbb{1}(t\in\{\bar{t}\})\log\left(\sigma\left(s^{e \to t}_{k,t}\right)\right) + \right.
\\
  & \left.(1-\mathbb{1}\left(t\in\{\bar{t}\}\right))\log\left(1 - \sigma\left(s^{e \to t}_{k,t}\right)\right)\right],
\end{aligned}
$
}
\end{equation}
where $\{\bar{t}\}$ is the set of text descriptions that describes enitity $k$, $\mathbb{1}(\cdot)$ is an indicator function testing if a text $t$ is within the descriptive texts $\{\bar{t}\}$ of the entity $k$. Finally, we combine the two-side contrast for entity-to-language alignment with:

\begin{equation} \label{loss:e2l}
\begin{aligned}
    \mathcal{L}_\text{e2l} = \alpha \frac{1}{N_\text{t}}\sum_{i=1}^{N_\text{t}}\ell^{t \to e}_i + \beta \frac{1}{K}\sum_{k=1}^{K}\ell^{e \to t}_{k},
\end{aligned}
\end{equation}
where factors $\alpha$ and $\beta$ serve to balance the scale of cross-entropy loss and binary cross-entropy loss.

\subsection{Model Learning}
\label{method:training_obj}

\paragraph{Training Data} We obtain training data following the data curation pipeline from SceneVerse~\cite{jia2024sceneverse}. Specifically, we use an off-the-shelf 3D instance segmentation model~\cite{zhu2024unifying} to generate entity mask proposals on ScanNet and generate referential and descriptive text of objects with the help of scene graphs and 2D foundation models~\cite{openai2023gpt4v} for fair comparisons with existing methods. We also leverage the generated paired \ac{3dvl} data from other real 3D scene datasets, including 3RScan~\cite{wald2019rio}, HM3D~\cite{ramakrishnan2021habitat} and MultiScan~\cite{mao2022multiscan} in SceneVerse as additional training data.

\vspace{-3mm}
\paragraph{Learning Objectives} Our training objective is fully contrastive-based. We combine both the point-to-entity alignment in~\cref{loss:e2e} and the entity-to-language alignment in~\cref{loss:e2l}, enabling the end-to-end training of the 3D encoder $\bm{\mathcal{E}}_{\text{pcd}}$ and the VL-adapter $\mathrm{Proj}_{\text{VL}}(\cdot)$, leading to a final loss function as:
\begin{equation}
\begin{aligned}
    \mathcal{L}_{\text{overall}} = \mathcal{L}_\text{p2e} + \mathcal{L}_\text{e2l}.
\end{aligned}
\vspace{-1mm}
\end{equation}
\paragraph{Implementation Details} For our default model, we adopt sparse-convolution based UNet (SPUNet)~\cite{graham20183d, choy20194d} as the 3D encoder and CLIP~\cite{radford2021learning} as the text encoder. The VL-adapter is implemented with a two-layer MLP. During training, we only update the parameters of the 3D encoder and VL-adapter and keep the text encoder frozen. We provide more training details in the \supp.
\section{Experiments}
\label{sec:exp}
In our experiments, we focus the following questions:
\begin{itemize}[leftmargin=*,nolistsep,noitemsep]
\item How effective are scene representations learned by \model for open-vocabulary scene understanding tasks?
\item Does the learned model show zero-shot transfer capabilities in data-scarce scenarios?
\item How do the learned semantically aligned representation compare with previous pre-training methods for diverse \ac{3dvl} tasks, including segmentation, object referral, question answering, \etc?
\item How does each component of \model contribute to model performance? Does training data domain and language type affect model performance?
\end{itemize}
In the following sections, we detail the design of experiments to address these key questions and demonstrate the effectiveness of \model as a general open-vocabulary pre-training pipeline as well as a plug-and-play backbone for various downstream tasks. We provide more experiment details and additional experiment results in the \supp.

\subsection{Open-vocabulary Scene Understanding}
\label{sec:exp:ov}
\begin{table}[htbp]
    \caption{\textbf{Open-Vocabulary 3D Semantic Segmentation on ScanNet}~\cite{dai2017scannet}. We report foreground mIoU (f-mIoU, \%) and foreground mAcc (f-mAcc, \%). $\dagger$ and $\ddagger$ mean results reproduced by RegionPLC~\cite{yang2024regionplc} and Uni3D~\cite{zhou2023uni3d}, independently.}
    \centering
    \resizebox{\linewidth}{!}{
    \label{tab:open_vocab_sem_seg_scannet}
        \begin{tabular}{l|c|cc}
            \toprule
            Method                     & Network              & f-mIoU                         & f-mAcc\\
            \midrule
            MaskCLIP$^{\dagger}$~\cite{zhou2022extract}      & CLIP~\cite{radford2021learning}                  & 23.1                           & 40.9     \\
            OpenScene-2D~\cite{peng2023openscene}               & LSeg~\cite{li2022language}                 & 58.0                           & 68.5        \\
            \midrule
            CLIP$^{2{\ddagger}}$~\cite{zeng2023clip2}      & CLIP~\cite{radford2021learning}                  & -                              & 38.5       \\
            Uni3D$^{{\ddagger}}$~\cite{zhou2023uni3d}      & ViT~\cite{dosovitskiy2020image}                   & -                              & 45.8      \\
            \midrule
            OpenScene-3D$^{\dagger}$~\cite{peng2023openscene}  & SPUNet16~\cite{graham20183d}          & 57.2                           & 69.9      \\
            OpenScene-3D$^{\dagger}$~\cite{peng2023openscene}  & SPUNet32~\cite{graham20183d}          & 57.8                           & 70.3      \\
            PLA$^{\dagger}$~\cite{ding2023pla}           & SPUNet16~\cite{graham20183d}          & 17.7                           & 33.5        \\
            PLA$^{\dagger}$~\cite{ding2023pla}           & SPUNet32~\cite{graham20183d}          & 19.1                           & 41.5        \\
            RegionPLC~\cite{yang2024regionplc}                  & SPUNet16~\cite{graham20183d}          & 56.9                           & 75.6         \\
            RegionPLC~\cite{yang2024regionplc}                  & SPUNet32~\cite{graham20183d}          & 59.6                           & 77.5         \\
            OV3D~\cite{jiang2024open}                       & SPUNet16~\cite{graham20183d}          & 64.0                           & 76.3     \\
            \midrule
            \textbf{\model}                       & SPUNet16~\cite{graham20183d}          & 64.6                           & 79.5      \\
            \textbf{\model}                       & SPUNet32~\cite{graham20183d}          & \textbf{66.0}                           & \textbf{81.3}      \\
            \bottomrule
        \end{tabular}
    }
    \vspace{-4mm}
    \label{tab:open_vocab_sem_seg}
\end{table}

\paragraph{Experiment Settings} We select \ac{ovsem} as a signature task for evaluating models' open-vocabulary scene understanding capabilities. Specifically, we follow previous works~\cite{ding2023pla, yang2024regionplc, jiang2024open} to evaluate the foreground mIoU (f-mIoU, \%) and foreground mAcc (f-mAcc, \%) on ScanNet by providing all class categories from ScanNet20 as input texts. We directly use the aligned 3D features $\mF_{\text{VL}}$ without applying the cross-view augmentation and pass the category texts into the frozen CLIP text encoder, using the best match between the two features as the prediction for the final class category. 
\vspace{-3mm}
\paragraph{Results \& Analyses} 
As shown in~\cref{tab:open_vocab_sem_seg_scannet}, \model achieves \sota performance, with 66.0\% f-mIoU and 81.3\% f-mAcc, outperforming previous \sota OV3D~\cite{jiang2024open} by 3.0\% and 6.5\%, respectively.

\model also shows a significant improvement ($\sim$10\%) over OpenScene, which uses pre-trained \ac{2dvl} model as supervision for 3D representation learning. This highlights the benefit of incorporating 3D information into scene representation learning. Additionally, we report model performance using the SPUNet16 architecture for the 3D encoder, demonstrating consistent improvements over previous methods with the same backbone. These results showcase better alignment of our learned 3D scene representations to language descriptions using the \model framework.
\vspace{-3mm}
\paragraph{Per-category Results}
Due to the page limit, we provide per-category performance analysis on ScanNet20~\cite{dai2017scannet} in the \textit{supplementary}. In conclusion, we find that \model achieves much higher IoU than previous open-source \sota RegionPLC~\cite{yang2024regionplc} on many categories with similar accuracies, indicating fewer false positives and better geometric understanding. 

\begin{figure}[t]
    \centering
    \includegraphics[width=\linewidth]{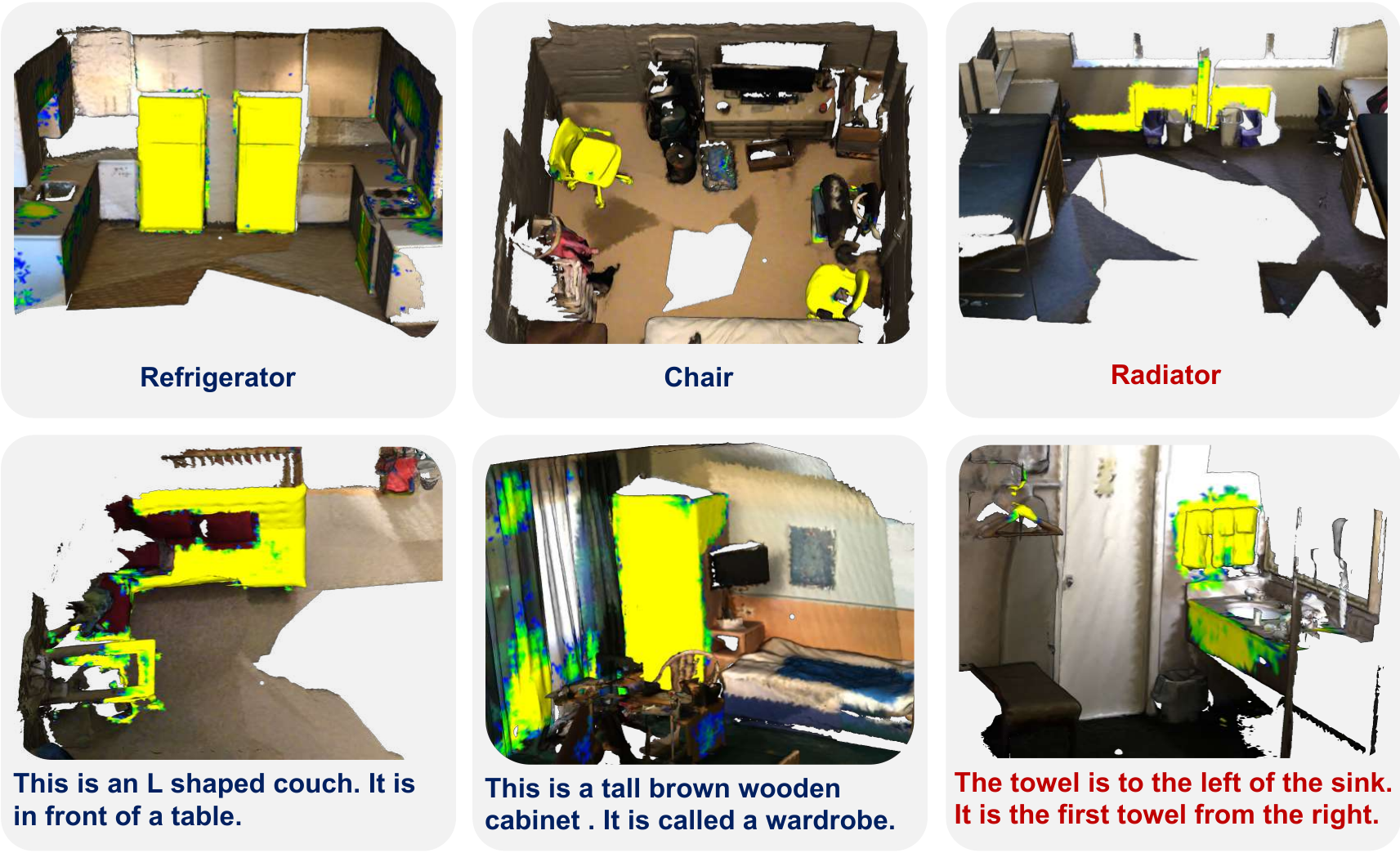}
    \caption{\textbf{Qualitative Results on ScanNet}~\cite{dai2017scannet}. We show relatively \textbf{good} and \textbf{bad} cases in \textcolor[RGB]{5, 34, 92}{\textbf{blue}} and \textcolor[RGB]{176, 35, 24}{\textbf{red}}.}
    \vspace{-4mm}
    \label{fig:qualitative_results}
\end{figure}

\vspace{-3mm}
\paragraph{Qualitative Results} 
We provide more qualitative results in~\cref{fig:qualitative_results}, including relatively \textcolor[RGB]{5, 34, 92}{\textbf{good}} and \textcolor[RGB]{176, 35, 24}{\textbf{bad}} cases. The basic object category understanding and more complex object referral results are shown in the top row and bottom row in~\cref{fig:qualitative_results}, respectively. \model shows superior basic category understanding capabilities with clear object boundaries, which attribute to the fine-grained geometry understanding. Moreover, \model can handle visual grounding tasks that take more complex texts as input. However, when handling distractions with similar appearances and locations, \model still faces challenges. We attribute this to the limitations of the fixed CLIP text encoder, which struggles to handle long and detailed descriptions required for precise visual grounding. We provide further investigation on this topic in the \textit{supplementary}.

\subsection{Zero-shot Transfer and Long-tail Scenarios}\label{sec:exp:data_eff}
\paragraph{Experiment Settings} To assess the generalization capability of learned representations to unseen scenes and language descriptions, we evaluate the f-mIoU and f-mAcc on \ac{ovsem} in two settings:

\begin{itemize}[leftmargin=*,nolistsep,noitemsep]

\item \textit{\textbf{Zero-shot Transfer}}: For \textit{zero-shot transfer} experiments, we evaluate \model on the unseen SceneVerse-val~\cite{jia2024sceneverse} and Matterport3D~\cite{chang2017matterport3d}. To clarify, the SceneVerse-val zero-shot transfer setting requires excluding MultiScan~\cite{mao2022multiscan} scenes and languages from training while testing on MultiScan data. We adhered to this protocol, removing MultiScan data during training for fair comparisons.
\item \textit{\textbf{Long-tail Scenarios}}: We provide the performance on ScanNet200~\cite{rozenberszki2022language} dataset, which is a more challenging scenario and includes a large number of test classes.
\end{itemize}
\vspace{-3mm}
\paragraph{Results \& Analyses} We present experimental results on \textit{zero-shot transfer} in~\cref{tab:open_vocab_sem_seg_zero_shot_ms} and ~\cref{tab:open_vocab_sem_seg_zero_shot_mp}, as well as \textit{long-tail scenarios} in~\cref{tab:open_vocab_sem_seg_long_tail_sn200}. Our findings are as follows:

\begin{itemize}[leftmargin=*,nolistsep,noitemsep]
    \item \textit{\textbf{Zero-shot Transfer}}: For \textit{zero-shot transfer} on SceneVerse-val~\cite{jia2024sceneverse}, as shown in~\cref{tab:open_vocab_sem_seg_zero_shot_ms}, \model significantly outperforms existing methods by a large margin. Specifically, for f-mIoU, we outperform the second-best OpenScene~\cite{peng2023openscene} by 3.7\%, and for f-mAcc, we surpass the second-best RegionPLC~\cite{yang2024regionplc} by 7.2\%. We further provide results on Matterport3D~\cite{chang2017matterport3d} in~\cref{tab:open_vocab_sem_seg_zero_shot_mp}. It's worth noting that, compared with OpenScene~\cite{peng2023openscene} and OV3D~\cite{jiang2024open}, our \model omits Matterport3D during training. However, \model achieves comparable performance with these domain-specific methods. Specifically, we achieve 47.7\% f-mIoU and 69.8\% f-mAcc. Compared with OV3D~\cite{jiang2024open}, \model is 2.7\% lower on f-mIoU, but significantly outperforms 4.1\% f-mAcc. \model also surpasses another \textit{zero-shot} method RegionPLC~\cite{yang2024regionplc} by about 20\% on both f-mIoU and f-mAcc. These results show that \model can generalize better to unseen scenarios.

    \begin{table}[t]
    \caption{\textbf{\textit{Zero-Shot} Open-Vocabulary 3D Semantic Segmentation on SceneVerse-val}~\cite{jia2024sceneverse}. We report f-mIoU (\%) and f-mAcc (\%).}
    \centering
        \label{tab:open_vocab_sem_seg_zero_shot_ms}
        \begin{tabular}{l|cc}
            \toprule
            Method & f-mIoU & f-mAcc \\
            \midrule
            OpenScene~\cite{peng2023openscene} & 41.3 & 52.5 \\
            RegionPLC~\cite{yang2024regionplc} & 39.1 & 56.4 \\
            \midrule
            \textbf{\model} & \textbf{45.0} & \textbf{63.6} \\
            \bottomrule
        \end{tabular}
\end{table}
    \begin{table}[t]
    \caption{\textbf{Open-Vocabulary 3D Semantic Segmentation on Matterport3D}~\cite{chang2017matterport3d}. We report the f-mIoU~(\%) and f-mAcc~(\%). $\ddagger$ means results reproduced by OV3D~\cite{jiang2024open}.}
    \centering
    \resizebox{0.9\linewidth}{!}{
        \begin{tabular}{l|c|cc}
        \toprule
        Method & \textit{Zero-Shot} & f-mIoU & f-mAcc \\
        \midrule
        OpenScene-3D$^{\ddagger}$~\cite{peng2023openscene} & \xmark & 49.7 & 64.0 \\
        OV3D~\cite{jiang2024open} & \xmark & \textbf{50.4} & 65.7 \\
        \midrule
        RegionPLC~\cite{yang2024regionplc} & \cmark & 28.9 & 43.8 \\
        \model & \cmark & 47.7 & \textbf{69.8} \\
        \bottomrule
        \end{tabular}
    }
    \vspace{-3mm}
    \label{tab:open_vocab_sem_seg_zero_shot_mp}
\end{table}

    \item \textit{{\textbf{Long-tail Scenarios}}}: We provide the results on ScanNet200~\cite{rozenberszki2022language} in~\cref{tab:open_vocab_sem_seg_long_tail_sn200}. \model achieves \sota performance, with 10.8\% f-mIoU and 27.4\% f-mAcc, outperforming existing models by a large margin ($\sim$10\% improvement in f-mAcc). This indicates that MPEC can capture fine-grained taxonomic distinctions, ultimately benefiting complex downstream embodied tasks.
\end{itemize}

\subsection{Fine-tuning on Downstream Tasks}\label{sec:exp:finetune}

\paragraph{Experiment Settings} We evaluate the adaptability of our learned 3D encoder as a general 3D backbone for fine-tuning on diverse 3D tasks. Specifically, we consider the following two levels of tasks:
\begin{itemize}[leftmargin=*,nolistsep,noitemsep]
    \item \textit{\textbf{Low-level Perception}}: For this group of tasks, we consider closed-set semantic and instance segmentation as representatives to compare our model against previous 3D point cloud representation learning methods. Concretely, we compare \model with unsupervised scene representation methods following the setting in~\cite{wang2024groupcontrast} on ScanNet and ScanNet200, using mIoU and mAP@0.5 as metrics for semantic and instance segmentation respectively. We fine-tune our learned 3D scene representation $\mF_P$ from the 3D encoder on the downstream tasks.
    We also conduct data efficiency experiments following~\cite{hou2021exploring}.
    
    \item \textit{\textbf{High-level Reasoning}}: We further test if the scene representation learned by \model can be directly adopted in more \ac{3dvl} tasks including grounding, captioning, and question answering. To achieve this goal, we select a recent \sota \ac{3dvl} model, PQ3D~\cite{zhu2024unifying}, that leverages the same backbone architecture as the ones used in \model for comparative studies. By replacing PQ3D's 3D encoder with our pre-trained method, we report model performance with fine-tuning on: (i) visual grounding datasets, including ScanRefer~\cite{chen2020scanrefer}, Nr3D and Sr3D~\cite{achlioptas2020referit3d}, Multi3dRefer~\cite{zhang2023multi3drefer}; (ii) 3D question-answering dataset SQA3D~\cite{ma2022sqa3d}; and (iii) scene captioning dataset Scan2Cap~\cite{chen2021scan2cap}. We report model performance under the convention of each dataset. Specifically, for grounding results on ScanRefer~\cite{chen2020scanrefer} and Multi3dRefer~\cite{zhang2023multi3drefer}, we report the grounding accuracies of predictions whose object bounding boxes overlap the ground truth with IoU > 0.5 (Acc@0.5). For visual grounding on Nr3D and Sr3D of Referit3D~\cite{achlioptas2020referit3d}, we report the grounding accuracies using ground-truth masks (Acc). For 3D dense captioning on Scan2Cap~\cite{chen2021scan2cap}, we report CIDEr under IoU@0.5 (CIDEr@0.5). For situated reasoning on SQA3D~\cite{ma2022sqa3d}, we report the answer accuracies (AccQA).
\end{itemize}

\vspace{-3mm}
\begin{table}[t]
    \caption{\textbf{Open-Vocabulary 3D Semantic Segmentation on ScanNet200}~\cite{rozenberszki2022language}.  We report f-mIoU (\%) and f-mAcc (\%). $\ddagger$ means results reproduced by OV3D~\cite{jiang2024open}.}
        \centering
        \begin{tabular}{l|cc}
            \toprule
            Method & f-mIoU & f-mAcc \\
            \midrule
            PLA~\cite{ding2023pla} & 1.8 & 3.1 \\
            OpenScene-3D$^{\ddagger}$~\cite{peng2023openscene} & 7.3 & - \\
            RegionPLC~\cite{yang2024regionplc} & 9.1 & 17.3 \\
            OV3D~\cite{jiang2024open} & 8.7 & - \\
            \midrule
            \textbf{\model} & \textbf{10.8} & \textbf{27.4} \\
            \bottomrule
        \end{tabular}
    \vspace{-3mm}
    \label{tab:open_vocab_sem_seg_long_tail_sn200}
\end{table}
\begin{table*}[ht]
\caption{\textbf{Fine-Tuning Results on Perception Tasks.} We report mIoU~(\%) for semantic segmentation and mAP@0.5~(\%) for instance segmentation on ScanNet~\cite{dai2017scannet} and ScanNet200~\cite{rozenberszki2022language}.}
    \centering
    \resizebox{\linewidth}{!}{
        \begin{tabular}{l| cccccc| cccccc}
            \toprule
            \multirow{2}[2]{*}{Datasets} & \multicolumn{6}{c|}{Semantic Segmentation (mIoU)}  & \multicolumn{6}{c}{Instance Segmentation (mAP@0.5)} \\ 
            \cmidrule(lr){2-7}\cmidrule(lr){8-13}
                                      & SC  & PC~\cite{xie2020pointcontrast}    & CSC~\cite{hou2021exploring}   & MSC~\cite{wu2023masked}   & GC~\cite{wang2024groupcontrast}   & \textbf{\model}  & SC     & PC~\cite{xie2020pointcontrast}     & CSC~\cite{hou2021exploring}    & MSC~\cite{wu2023masked}   & GC~\cite{wang2024groupcontrast}    & \textbf{\model}   \\ 
            \midrule
            ScanNet~\cite{dai2017scannet}                   & 72.2    & 74.1             & 73.8              & 75.3  & 75.7            & \textbf{75.8}   & 56.9   & 58.0   & 59.4   & 59.6  & 62.3  & \textbf{62.5} \\
            ScanNet200~\cite{rozenberszki2022language}                & 25.0                            & 26.2             & 26.4              & 28.8                            & 30.0          & \textbf{31.8}       & 24.5   & 24.9   & 25.2   & 26.8  & 27.5  & \textbf{31.6}   \\
            \bottomrule
            \end{tabular}
        }
        \label{tab:finetune_exp}
\vspace{-3mm}
\end{table*}
\begin{table}[ht]
    \caption{\textbf{Fine-Tuning Results on Reasoning Tasks.} The grounding accuracy of ScanRefer~\cite{chen2020scanrefer} and Multi3DRefer~\cite{zhang2023multi3drefer} are reported under IoU@0.5. The results of Nr3D and Sr3D~\cite{achlioptas2020referit3d} are reported using ground-truth masks. We report CIDEr under IoU@0.5 on Scan2Cap~\cite{chen2021scan2cap} and answer accuracy on SQA3D~\cite{ma2022sqa3d}.} 
    \centering
    \resizebox{\linewidth}{!}{
        \label{tab:reasoning_tasks}
        \begin{tabular}{l|cccccc}
        \toprule
        {\multirow{2}{*}{Method}}  & ScanRefer & Nr3D & Sr3D & Multi3DRefer & Scan2Cap  & SQA3D \\
                                & Acc@0.5   & Acc  & Acc  & F1@0.5       & CIDEr@0.5 & AccQA   \\ 
        \midrule
        3DJCG                                       & 37.3      & -    & -    & 26.6         & 47.7      & -     \\
        M3DRef-CLIP                                 & 44.7      & 49.4 & -    & 38.4         & -         & -     \\
        3D-VisTA                                    & 45.8      & 64.2 & 76.4 & -            & 66.9      & \textbf{48.5}  \\
        Vote2CapDetr                                & -         & -    & -    & -            & 64.3      & -     \\
        \midrule
        PQ3D                                        & 51.2      & \textbf{66.7} & 79.7 & 50.1         & \textbf{80.3}      & 47.1  \\
        \model + PQ3D                          & \textbf{51.8}      & \textbf{66.7} & \textbf{80.0} & \textbf{50.8}         & 80.2      & 47.5  \\ 
        \bottomrule
        \end{tabular}
    }
\vspace{-3mm}
\end{table}

\paragraph{Results \& Analyses} We provide fine-tuning results for low-level perception tasks in~\cref{tab:finetune_exp} and results for high-level reasoning tasks in~\cref{tab:reasoning_tasks}. Our findings on these two groups of tasks are as follows:
\begin{itemize}[leftmargin=*,nolistsep, noitemsep]
    \item \textit{\textbf{Low-level Perception}}: For close-set semantic segmentation, \model achieves considerable performance gains over existing methods. Specifically, it surpasses previous methods on both ScanNet~\cite{dai2017scannet} and ScanNet200~\cite{rozenberszki2022language}, achieving mIoUs of 75.8\% and 31.8\%, respectively, with a notable 1.8\% improvement on ScanNet200 over GC~\cite{wang2024groupcontrast}. For close-set instance segmentation, \model performs comparably well to GC on ScanNet~\cite{dai2017scannet}, yet significantly outperforms on ScaNet200~\cite{rozenberszki2022language}. Specifically, our \model achieves mAP@0.5 of 31.6\%, achieving substantial improvements over the previous \sota GC's 27.5\%. We attribute these gains to our proposed masked point-entity contrastive learning framework, which provides more discriminative features across 3D entities. Data efficiency experiments are provided in \textit{supplementary}.

    \item \textit{\textbf{High-level Reasoning}}: As shown in~\cref{tab:reasoning_tasks}, when substituting the 3D backbone with \model for fine-tuning on high-level \ac{3dvl} tasks, the model achieves consistent and notable improvements in most reasoning tasks. Specifically, \model with PQ3D task heads sets new \sota performance on 3D visual grounding datasets, including ScanRefer~\cite{chen2020scanrefer}, Nr3D~\cite{achlioptas2020referit3d}, Sr3D~\cite{achlioptas2020referit3d}, and Multi3dRefer~\cite{zhang2023multi3drefer}. Additionally, our model also boosts the accuracy on SQA3D~\cite{ma2022sqa3d} from 47.1\% to 47.5\%. These findings highlight \model’s adaptability and its potential to serve as an initial backbone weight for various downstream tasks.
\end{itemize}

\subsection{Additional Discussions}
\label{sec:exp:ablation}

In this subsection, we conduct ablation studies on model design and data engineering.

\vspace{-3mm}
\paragraph{Model Design}
In~\cref{tab:model_ablation}, we show the results with different model designs on open-vocabulary 3D semantic segmentation on ScanNet~\cite{dai2017scannet}. Specifically, directly applying $\mathcal{L}_{\text{e2e}}$ with $\mathcal{L}_{\text{e2l}}$ without employing cross mask augmentations leads to a noticeable drop on f-mIoU and f-mAcc, compared to minimizing $\mathcal{L}_{\text{e2l}}$ alone. We hypothesize this may be attributed to the model's tendency to overemphasize the uniqueness of each entity within a single view, thereby neglecting shared attributes across entities. By introducing masked cross-view augmentations, minimizing $\mathcal{L}_{\text{e2e}}$ across views encourages the model to distinguish entities between views without enforcing distinctiveness within the same view. This adjustment helps preserve common attributes for different entities, such as semantic categories.

\vspace{-3mm}
\paragraph{Text Descriptions Type}
The text descriptions of entities include two types: captions $T_\text{cap}$ which provide detailed descriptions of an object's visual and physical properties, and referrals $T_\text{ref}$ which articulate the spatial relationships between the entity and other objects within the scene. We conduct an ablation study on each description type in~\cref{tab:lang_type_ablation}. As shown in~\cref{tab:lang_type_ablation}, leveraging both description types yields the best performance. This result aligns with our expectations, as combining an entity's inherent attributes with its relationships with other entities provides a more comprehensive understanding of the 3D entity.

\begin{table}[tbp]
    \caption{\textbf{Ablation Analysis on Model Design.}  We report f-mIoU~(\%) and f-mAcc~(\%) on ScanNet~\cite{dai2017scannet}.}
    \centering
    \begin{tabular}{ccc|cc}
         \toprule
         $\mathcal{L}_{\text{e2l}}$ & $\mathcal{L}_{\text{e2e}}$ & Cross-View Aug. & f-mIoU & f-mAcc \\
         \midrule
         {\cmark} &  &  & 63.6 & 79.2 \\
         {\cmark} & {\cmark} &  & 62.2 & 78.5 \\
         {\cmark} & {\cmark} & {\cmark} & \textbf{64.6} & \textbf{79.5} \\
         \bottomrule
    \end{tabular}
    \label{tab:model_ablation}
\end{table}

\begin{table}[htbp]
    \caption{\textbf{Ablation Analysis on Text Type.} We report f-mIoU~(\%) and f-mAcc~(\%) on ScanNet~\cite{dai2017scannet}.}
    \centering
    \begin{tabular}{cc|cc}
        \toprule
         $T_{\text{cap}}$ & $T_{\text{ref}}$ & f-mIoU & f-mAcc \\
         \midrule
         {\cmark} &  & 57.6 & 74.8 \\
          & {\cmark} & 61.7 & 77.2 \\
         {\cmark} & {\cmark} & \textbf{64.6} & \textbf{79.5} \\
         \bottomrule
    \end{tabular}
    \vspace{-4mm}
    \label{tab:lang_type_ablation}
\end{table}

\vspace{-3mm}
\paragraph{Training Dataset}
Our training dataset, SceneVerse~\cite{jia2024sceneverse}, is composed of seven datasets, including real-world scene datasets ScanNet~\cite{dai2017scannet}, MultiScan~\cite{mao2022multiscan}, RScan~\cite{wald2019rio}, ARKitScenes~\cite{baruch2021arkitscenes} and HM3D~\cite{ramakrishnan2021habitat} as well as synthetic environments from Structured3D~\cite{zheng2020structured3d} and ProcTHOR~\cite{NEURIPS2022_27c546ab}. For our ablation study, we discard ARKitScenes~\cite{baruch2021arkitscenes} and the synthetic datasets, \ie Structured3D~\cite{zheng2020structured3d} and ProcTHOR~\cite{NEURIPS2022_27c546ab}, to analyze the influence of different real-world data sources, as shown in~\cref{tab:data_scale_ablation}. We observe obvious performance gains when scaling up the training data.

\begin{table}[htbp]
    \caption{\textbf{Ablation Analysis on Training Dataset.} We report f-mIoU and f-mAcc on ScanNet~\cite{dai2017scannet} after training on different combinations of datasets.}
    \centering
    \resizebox{1.0\linewidth}{!}{
        \begin{tabular}{cccc|cc}
            \toprule
             ScanNet~\cite{dai2017scannet} & MultiScan~\cite{mao2022multiscan} & RScan~\cite{wald2019rio} & HM3D~\cite{ramakrishnan2021habitat} & f-mIoU & f-mAcc \\
             \midrule
             {\cmark} &  &  &  & 56.4 & 74.3 \\
             {\cmark} & {\cmark} &  &  & 57.7 & 75.1 \\
             {\cmark} &  & {\cmark} &  & 59.9 & 76.5 \\
             {\cmark} &  &  & {\cmark} & 63.3 & 79.3 \\
             {\cmark} & {\cmark} & {\cmark} &  & 60.6 & 77.4 \\
             {\cmark} & {\cmark} &  & {\cmark} & 63.3 & 79.0 \\
             {\cmark} &  & {\cmark} & {\cmark} & 62.7 & 78.3 \\
             {\cmark} & {\cmark} & {\cmark} & {\cmark} & \textbf{64.6} & \textbf{79.5} \\
             \bottomrule
        \end{tabular}
    }
    \vspace{-3mm}
    \label{tab:data_scale_ablation}
\end{table}
\section{Conclusion}
\label{sec:conclusion}
In this paper, we propose a novel entity-level contrastive learning framework, \model, for open-vocabulary 3D scene understanding, which achieves \sota performance on open-vocabulary 3D semantic segmentation. Through fine-tuning on diverse downstream tasks, our model demonstrates high effectiveness and transferability, showcasing its potential as a foundational approach for 3D vision-language applications. Furthermore, our data scale ablation study reveals the importance of large-scale 3D vision-language datasets in advancing 3D foundation models, underscoring a critical area for future development.

\clearpage
{
    \small
    \bibliographystyle{ieeenat_fullname}
    \bibliography{main}
}

\clearpage

\appendix

\renewcommand{\thefigure}{A.\arabic{figure}}
\renewcommand{\thetable}{A.\arabic{table}}
\renewcommand{\theequation}{A.\arabic{equation}}

\maketitlesupplementary
\setcounter{table}{0}

\section{Implementation Details}
We provide implementation and training details of our proposed \textit{Masked Point-Entity Contrast} (MPEC) model.

\noindent\paragraph{Model Architecture} For open-vocabulary 3D semantic segmentation and zero-shot visual grounding experiments, we ensure a fair comparison with existing methods by employing SparseUNet16 and SparseUNet32~\cite{graham20183d} as the 3D encoder and the frozen CLIP~\cite{radford2021learning} as the text encoder. The vision-language adapter consists of a two-layer MLP. For downstream fine-tuning tasks, we train a MinkUNet34C~\cite{choy20194d} re-implemented with \textit{SpConv}~\cite{spconv2022} following \textit{Pointcept}~\cite{pointcept2023} using the same point-entity contrastive supervision. 

For fine-tuning on low-level perception tasks, we build our experiments using the point cloud perception codebase \textit{Pointcept}~\cite{pointcept2023}. Specifically, for all the semantic segmentation tasks, including closed-set experiments on ScanNet~\cite{dai2017scannet}, ScanNet200~\cite{rozenberszki2022language}
, we add a single linear layer as the classification head and utilize the cross-entropy loss for supervision. For instance segmentation tasks, we follow previous works~\cite{xie2020pointcontrast, hou2021exploring, wu2023masked, wang2024groupcontrast} to adopt PointGroup~\cite{jiang2020pointgroup} as the instance segmentation head.

For fine-tuning high-level reasoning tasks, we select PQ3D~\cite{zhu2024unifying}, a \sota framework for reasoning tasks for indoor scenes as the baseline. Specifically, we replace the voxel encoder of PQ3D with our MinkUNet34C while maintaining other configurations, and fine-tune on the reasoning datasets.

\noindent\paragraph{View Generation and Masking Strategy} We follow the view generation pipeline and mask strategy in MSC~\cite{wu2023masked}. For a given 3D point cloud, we first create two copies and apply separate random augmentation sequences to each, generating two distinct views of the same scene. The augmentation sequence, detailed in \cref{tab:supp_view_generation_pipeline}, consists of three main components: \colorbox[RGB]{229, 239, 219}{spatial augmentations}, \colorbox[RGB]{215, 234, 233}{photometric augmentations} and \colorbox[RGB]{243, 225, 180}{sampling augmentations}.

For the masking strategy, we adopt the approach proposed in MSC~\cite{wu2023masked}, setting the grid size to 0.1m to partition the original coordinates into evenly spaced, non-overlapping grids. For each view, 40\% of the grids are selected and the features inside are masked and replaced with learnable tokens. Importantly, the selected grids for the two views are mutually exclusive, ensuring no overlap.

\begin{table}[ht]
    \centering
    \caption{\textbf{View Generation Pipeline.}}
    \resizebox{\linewidth}{!}{
        \begin{tabular}{l|c}
            \toprule
            Augmentation & Value \\
            \midrule
            \rowcolor[RGB]{229, 239, 219} random rotate & angle=[-1/64, 1/64], axis=`x', p=1 \\
            \rowcolor[RGB]{229, 239, 219} random rotate & angle=[-1/64, 1/64], axis=`y', p=1 \\
            \rowcolor[RGB]{229, 239, 219} random flip & p=0.5 \\
            \midrule
            \rowcolor[RGB]{215, 234, 233} random coord jitter & sigma=0.005, clip=0.02 \\
            \rowcolor[RGB]{215, 234, 233} random color brightness jitter & ratio=0.4, p=0.8 \\
            \rowcolor[RGB]{215, 234, 233} random color contrast jitter & ratio=0.4, p=0.8 \\
            \rowcolor[RGB]{215, 234, 233} random color saturation jitter & ratio=0.2, p=0.8\\
            \rowcolor[RGB]{215, 234, 233} random color hue jitter & ratio=0.02, p=0.8 \\
            \rowcolor[RGB]{215, 234, 233} random color gaussian jitter & std=0.05, p=0.95 \\
            \midrule
            \rowcolor[RGB]{243, 225, 180} grid sample & grid size=0.02 \\
            \rowcolor[RGB]{243, 225, 180} random crop & ratio=0.6 \\
            center shift & n/a \\
            color normalize & n/a \\
            \bottomrule
        \end{tabular}
    }
    \label{tab:supp_view_generation_pipeline}
\end{table}
\noindent\paragraph{Training} For the point-entity contrastive learning, we utilize the AdamW optimizer with a learning rate of $1\times10^{-3}$ for 500 epochs with a cosine warm-up period of 200 steps. During training, we set a batch size of 4 scenes for each GPU and sample 64 text descriptions for each scene. To balance the scale of cross-entropy loss and binary cross-entropy loss in $\mathcal{L}_\text{e2l}$, we empirically set $\alpha$ and $\beta$ to 1.0 and 6.0, respectively. All the contrastive learning experiments are performed on 4 NVIDIA-A100 GPUs with the longest training taking less than 4 days.

\begin{table}
    \centering
    \caption{\textbf{Fine-Tuning Setting on Low-Level Perception Tasks.}}
    \begin{tabular}{l|c}
        \toprule
        Config & Value \\
        \midrule
        optimizer & SGD \\
        scheduler & cosine decay \\
        weight decay & 1e-4 \\
        optimizer momentum & 0.9 \\
        batch size & 12 \\
        warmup epochs & 40\\
        epochs & 800 \\
        \bottomrule
    \end{tabular}
    \label{tab:supp_low_level_finetuning_setting}
\end{table}
For the downstream fine-tuning experiments, we conduct all the low-level perception tasks on \textit{Pointcept}~\cite{pointcept2023} and all the high-level reasoning tasks on PQ3D~\cite{zhu2024unifying}. 

The general fine-tuning setting for low-level perception tasks is shown in \cref{tab:supp_low_level_finetuning_setting}. We adjust the learning rate based on the task. Specifically, for full-set semantic and instance segmentation fine-tuning experiments on ScanNet and ScanNet200, the learning rate is set to 0.2. 

\begin{table*}[t!]
\caption{\textbf{Partial Per-Category Performance on ScanNet}~\cite{dai2017scannet}. We compare the IoU~(\%) and accuracy~(\%) with previous SOTA RegionPLC~\cite{yang2024regionplc} of each category.}
\centering
    \begin{tabular}{l|cc|cc|cc|cc|cc|cc|cc}
          \toprule
          & \multicolumn{2}{c|}{chair} & \multicolumn{2}{c|}{bookshelf} & \multicolumn{2}{c|}{counter} & \multicolumn{2}{c|}{toilet} & \multicolumn{2}{c|}{sink} & \multicolumn{2}{c|}{shower curtain} & \multicolumn{2}{c}{curtain} \\
          & IoU         & Acc         & IoU           & Acc           & IoU          & Acc          & IoU          & Acc          & IoU              & Acc             & IoU          & Acc         & IoU         & Acc        \\
          \midrule
RegionPLC & 75.4        & 82.9        & 72.9          & \textbf{96.3}          & 49.0         & 64.6         & 64.2         & \textbf{98.6}        & 38.1        & 84.2       & 43.3             & \textbf{90.7}            & 46.5         & 51.4         \\
\midrule
MPEC      & \textbf{83.8}        & \textbf{85.5}        & \textbf{80.3}          & 92.3          & \textbf{56.4}         & \textbf{65.2}         & \textbf{85.6}         & 98.2        & \textbf{48.5}        & \textbf{85.6}       & \textbf{61.6}             & 87.1            & \textbf{66.1}         & \textbf{73.1}         \\
\bottomrule
\end{tabular}
\label{tab:scannet_per_category_analysis_full}
\end{table*}

For training PQ3D on high-level reasoning tasks, we train the model on multiple 3D vision-language tasks including visual grounding, question answering, and dense captioning for 50 epochs. The model architecture uses a hidden dimension of 768 and 4 query decoder layers. Optimization is performed using the AdamW optimizer with a learning rate of $1\times10^{-4}$, a batch size of 16, and momentum parameters $\beta_1 = 0.9$ and $\beta_2 = 0.98$. The loss balancing weights are set to $\lambda_{\mathrm{gen}} = 1$ and $\lambda_{\mathrm{grd}} = 10$.

\section{Additional Per-category Performance Analyses on ScanNet}
We provide part of the per-category performance on ScanNet in~\cref{tab:scannet_per_category_analysis_full}. As can be seen, though RegionPLC~\cite{yang2024regionplc} and \model achieve similar accuracies on many categories, \model continuously outperforms RegionPLC on the IoU metric by a large margin, indicating fewer false positives and better geometric understanding. Superior results on \textit{shower curtain} and \textit{curtain} further highlight \model's strong spatial reasoning and semantic understanding ability.

\section{Additional Experiment Results for \textit{Zero-shot} 3D Visual Grounding} 
In Fig.~3 of the main paper, we observe that \model still faces challenges when dealing with complicated grounding texts qualitatively. We attribute this phenomenon to the limitations of the fixed CLIP~\cite{radford2021learning} text encoder. This section provides quantitative analysis on \textit{zero-shot} 3D visual grounding experiments on SceneVerse~\cite{jia2024sceneverse} to support this hypothesis.

\paragraph{Experiment Settings}
Following the SceneVerse-val \textit{zero-shot} setting in~\cite{jia2024sceneverse}, we remove MultiScan data during training for fair comparisons. We provide the model ground-truth object proposals and use the pooled feature $\mF_\text{VL}$ for each object to match with the grounding text for predictions. We test the grounding accuracy of different text encoders, \ie, frozen CLIP~\cite{radford2021learning} and trainable BERT~\cite{devlin2018bert}.
\begin{table}[h]
    \caption{\textbf{\textit{Zero-Shot} 3D Visual Grounding on SceneVerse-val}~\cite{jia2024sceneverse}. We report accuracy~(\%) on SceneVerse-val~\cite{jia2024sceneverse} and evaluate models using GT object proposals. \textcolor{red}{\faFire} and \textcolor[RGB]{118, 197, 242}{\faSnowflake} indicates trainable and frozen text encoder, respectively.}
    \centering
    \resizebox{\linewidth}{!}{
        \begin{tabular}{l|c|ccc}
            \toprule
            Method & Text Encoder & Overall & Easy & Hard \\
            \midrule
            3D-VisTA~\cite{zhu20233d} & BERT$^{\text{\textcolor{red}{\faFire}}}$ & 52.9 & 59.6 & 35.4 \\
            \textbf{GPS}~\cite{jia2024sceneverse} & BERT$^{\text{\textcolor{red}{\faFire}}}$ & \textbf{59.2} & \textbf{69.4} & \textbf{44.0} \\
            \midrule
            OpenScene~\cite{peng2023openscene} & CLIP$^{\text{\textcolor[RGB]{118, 197, 242}{\faSnowflake}}}$ & 13.3 & 15.5 & 10.1 \\
            RegionPLC~\cite{yang2024regionplc} & CLIP$^{\text{\textcolor[RGB]{118, 197, 242}{\faSnowflake}}}$ & 10.6 & 11.8 & 8.9 \\
            \model & CLIP$^{\text{\textcolor[RGB]{118, 197, 242}{\faSnowflake}}}$ & 17.0 & 23.8 & 6.7 \\
            \model & BERT$^{\text{\textcolor{red}{\faFire}}}$ & 42.6 & 56.2 & 22.2 \\
            \bottomrule
        \end{tabular}
    }
    \label{tab:supp_refer_zero_shot_ms}
\end{table}

\paragraph{Results \& Analyses} 
We present experiments for \textit{zero-shot} 3D visual grounding on SceneVerse-val~\cite{jia2024sceneverse} in~\cref{tab:supp_refer_zero_shot_ms}. \model with the frozen CLIP text encoder achieves a better overall grounding accuracy of 17\% compared with existing available open-vocabulary 3D understanding models like OpenScene~\cite{peng2023openscene} and RegionPLC~\cite{yang2024regionplc}. However, compared with task-specific models for 3D visual grounding, \ie, 3D-VisTA~\cite{zhu20233d} and GPS~\cite{jia2024sceneverse}, \model with the frozen~(\textcolor[RGB]{118, 197, 242}{\faSnowflake}) CLIP text encoder is considerably lower by more than~35\%~(17$\%$ \vvss 52+\%). After replacing the frozen CLIP text encoder with a trainable~(\textcolor{red}{\faFire}) BERT, the overall accuracy significantly improves from 17\% to 42.6\%. This underscores the limitation of the frozen LCIP text encoder, which struggles to handle long and detailed descriptions, particularly when grounding specific 3D objects in complex 3D scenes.

\section{Additional Experiment Results for Data-efficiency Fine-tuning}
In this section, we provide additional fine-tuning experiment results on the ScanNet Data-Efficiency benchmark~\cite{hou2021exploring}. 

\paragraph{Experiment Settings} We compare our method with previous methods on ScanNet-LR~(Limited Scene Reconstruction) and ScanNet-LA~(Limited Annotation) test splits. For ScanNet-LR, we use the $\{1\%, 5\%, 10\%, 20\%\}$ sampled scenes provided in ScanNet-LR and use the annotations within each scene to fine-tune our pre-trained representation $\mF_{P}$ for semantic segmentation. Similarly, For ScanNet-LA, we follow~\cite{hou2021exploring} and provide $\{20, 50, 100, 200\}$ labeled points per scene for fine-tuning our learned representation. Notably, we train \model by removing ScanNet data under this setting and report the mIoU for semantic segmentation on both splits as the evaluation metric.
\begin{table*}[ht]
    \begin{minipage}[t]{0.495\textwidth}
    \caption{\textbf{ScanNet Limited Scene Resconstruction.} We report the mIoU~(\%) results on ScanNet~\cite{dai2017scannet} data efficient semantic segmentation benchmark with limited scene reconstruction setting. }
    \centering
    \resizebox{\linewidth}{!}{
        \begin{tabular}{c|ccccc}
        \toprule
        LR                        & \multicolumn{5}{c}{Semantic Segmentation (mIoU)} \\ 
        \midrule
        Pct.    & SC    & CSC~\cite{hou2021exploring}   & MSC~\cite{wu2023masked}   & GC~\cite{wang2024groupcontrast}   & Ours  \\ 
        \midrule
        1\%                       & 26.1        & 28.9              & 29.2                  & 30.7                          & \textbf{40.8}         \\
        5\%                       & 47.8        & 49.8              & 50.7                  & 52.9                          & \textbf{58.5}         \\
        10\%                      & 56.7        & 59.4              & 61.0                  & 62.0                          & \textbf{64.0}         \\
        20\%                      & 62.9        & 64.6              & 64.9                  & \textbf{66.5}                          & 66.3         \\
        \bottomrule
        \end{tabular}
    }
    \label{tab:lr_sem_seg}
    \end{minipage}
    \hfill
    \begin{minipage}[t]{0.489\textwidth}
        \caption{\textbf{ScanNet Limited Annotation.} We report the mIoU~(\%) results on ScanNet~\cite{dai2017scannet} data efficient semantic segmentation benchmark with limited point annotation setting.}
        \centering
        \resizebox{\linewidth}{!}{
            \begin{tabular}{c|ccccc}
            \toprule
            LA                        & \multicolumn{5}{c}{Semantic Segmentation (mIoU)} \\ 
            \midrule
            Pts.    & SC    & CSC~\cite{hou2021exploring}   & MSC~\cite{wu2023masked}   & GC~\cite{wang2024groupcontrast}   & Ours      \\
            \midrule
            20                       & 41.9      & 55.5              & 61.2                  & 61.2                          & \textbf{62.9}          \\
            50                       & 53.9      & 60.5              & 66.8                  & 67.3                          & \textbf{69.2}          \\
            100                      & 62.2        & 65.9              & 69.7                  & 70.3                          & \textbf{72.0}        \\
            200                      & 65.5        & 68.2              & 70.7                  & 71.8                          & \textbf{73.1}\\
            \bottomrule
            \end{tabular}
        }
        \label{tab:la_sem_seg}
    \end{minipage}
\end{table*}

\paragraph{Results \& Analyses}
As shown in~\cref{tab:lr_sem_seg} and~\cref{tab:la_sem_seg}, our method consistently outperforms previous methods by a large margin, particularly in scenarios with extremely limited reconstructions ($\sim$10\% improvement for 1\% trained scenes). This highlights the ability of \model to retain language-aligned 3D feature extraction on unseen scenes and the fast adaptability of the learned representations to downstream tasks under data-scarce scenarios.

\begin{figure*}[t]
    \centering
    \includegraphics[width=\linewidth]{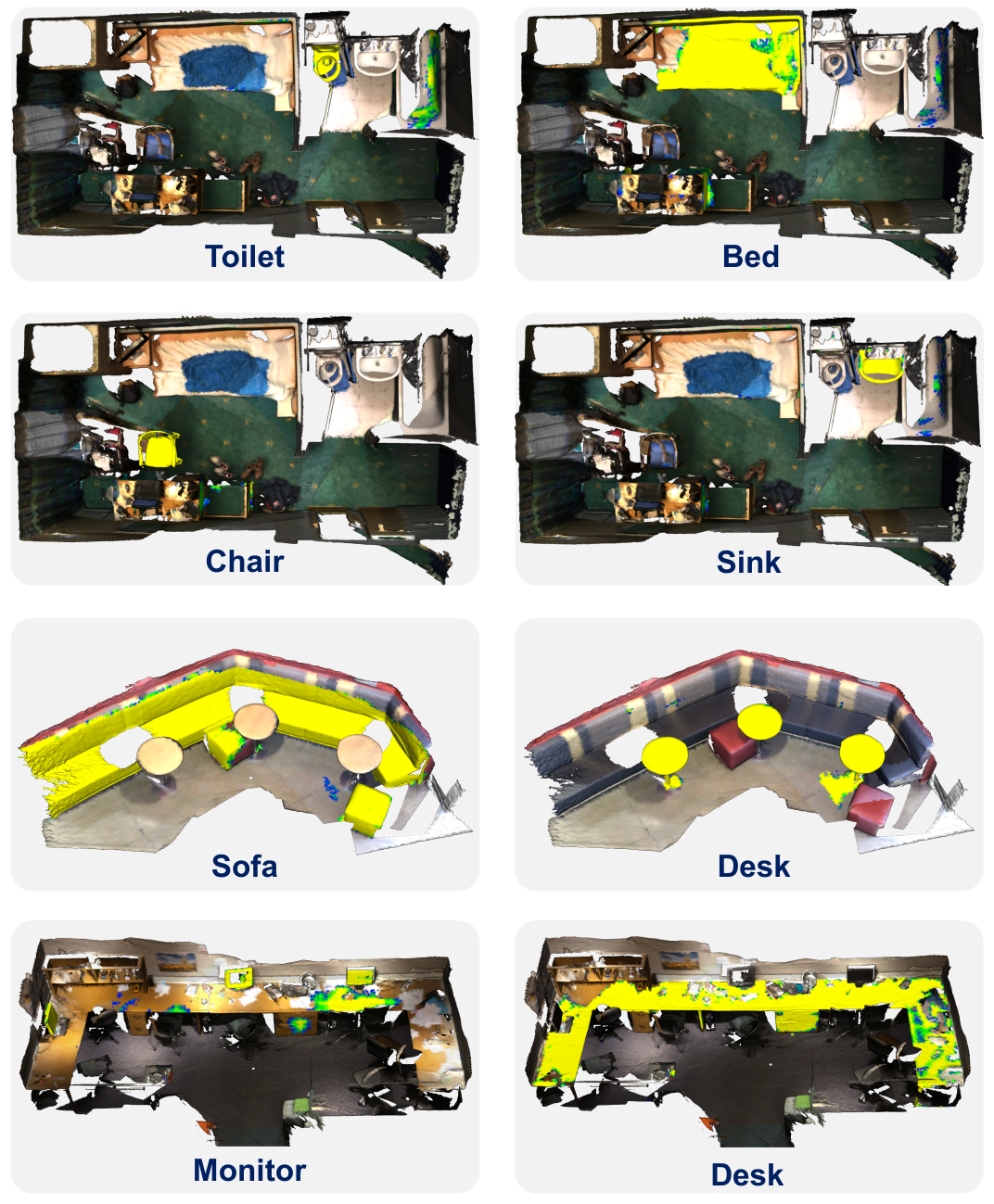}
    \caption{\textbf{More Qualitative Results on ScanNet}~\cite{dai2017scannet}.}
    \label{fig:supp_vis_1}
\end{figure*}

\begin{figure*}[t]
    \centering
    \includegraphics[width=\linewidth]{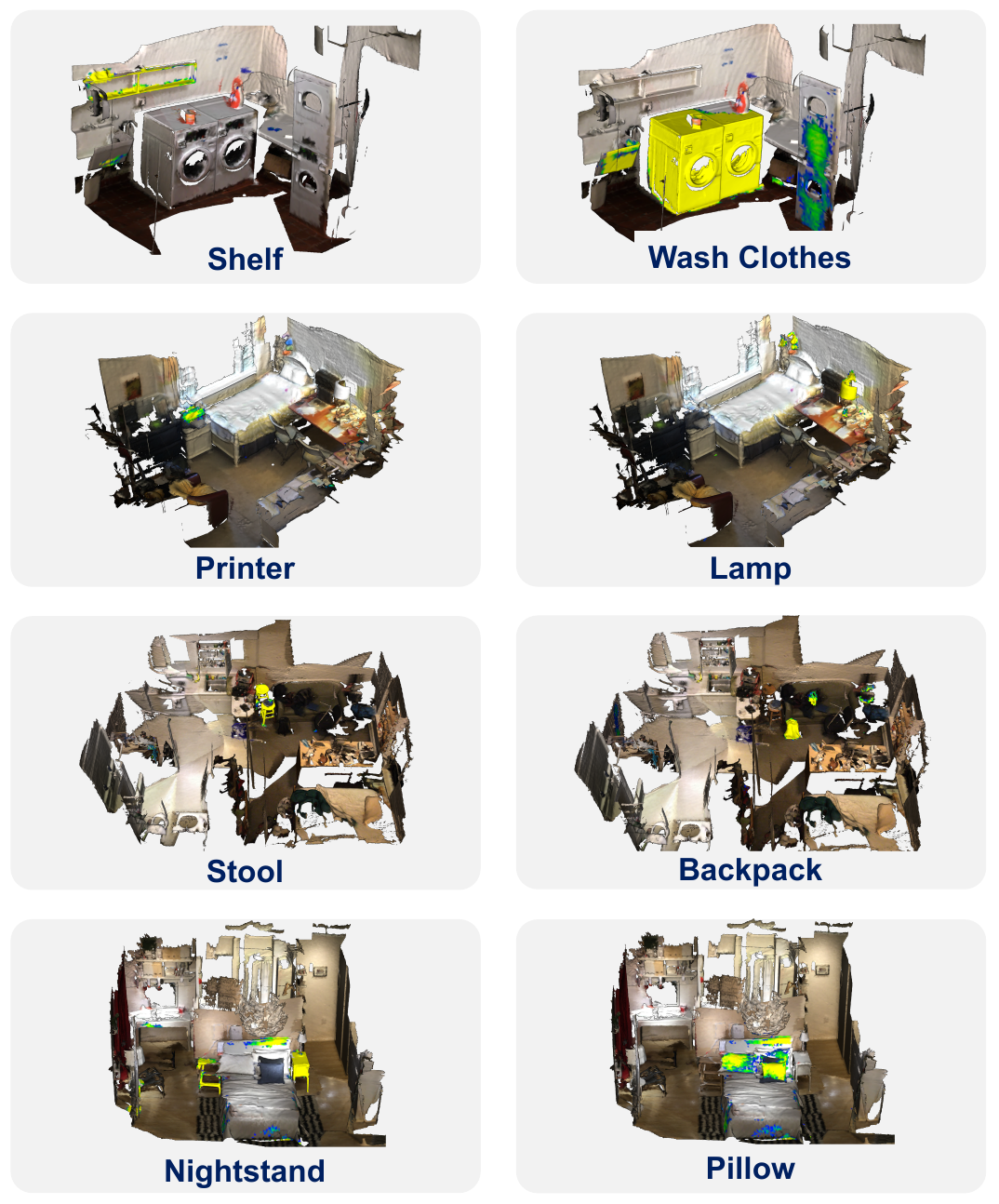}
    \caption{\textbf{More Qualitative Results on ScanNet}~\cite{dai2017scannet}.}
    \label{fig:supp_vis_2}
\end{figure*}

\section{More Quantitative Results}
We provide more qualitative results in~\cref{fig:supp_vis_1} and~\cref{fig:supp_vis_2}. 

\end{document}